\documentclass[sigconf]{acmart}

\begin{document}

\newcommand\todo[1]{\textcolor{red}{#1}}
\title{Unifying data for fine-grained visual species classification}

\author{Sayali Kulkarni}
\email{sayali@google.com}
\affiliation{%
  \institution{Google Research}
}

\author{Tomer Gadot}
\email{tomerg@google.com}
\affiliation{%
  \institution{Google Inc.}
}

\author{Chen Luo}
\email{chenluo@google.com}
\affiliation{%
  \institution{Google Inc.}
}

\author{Tanya Birch}
\email{tanyak@google.com}
\affiliation{%
  \institution{Google Earth Outreach}
  \institution{Wildlife Insights}
}

\author{Eric Fegraus}
\email{efegraus@conservation.org}
\affiliation{%
  \institution{Conservation International}
  \institution{Wildlife Insights}
}
%
\renewcommand{\shortauthors}{}

\acmConference[KDD'19]{Data Mining and AI for Conservation}{August 2019}{Anchorage, Alaska - USA}
\acmPrice{}
\acmDOI{}
\acmISBN{}

\begin{abstract}

Wildlife monitoring is crucial to nature conservation and has been done by manual observations from motion-triggered camera traps deployed in the field. Widespread adoption of such in-situ sensors has resulted in unprecedented data volumes being collected over the last decade. A significant challenge exists to process and reliably identify what is in these images efficiently. Advances in computer vision are poised to provide effective solutions with custom AI models built to automatically identify images of interest and label the species in them. Here we outline the data unification effort for the Wildlife Insights platform from various conservation partners, and the challenges involved. Then we present an initial deep convolutional neural network model, trained on 2.9M images across 465 fine-grained species, with a goal to reduce the load on human experts to classify species in images manually. The long-term goal is to enable scientists to make conservation recommendations from near real-time analysis of species abundance and population health.

\end{abstract}


\maketitle

\section{Introduction}
Wildlife Insights (WI) is a partnership of conservation organizations combining forces to advance our understanding of global species population health. WI is building a cloud-based platform that provides data management, processing, and analytical solutions available to any individual or organization. Central to this platform is the use of AI, mainly fine-grained species identification, to process camera trap images and transform them into metadata that can be used in analytics and reports for decision-makers. The WI partnership will bring together an unprecedented volume of images (20M just with core partners) and diversity of images (in terms of geographic distribution, variability in sensors used, temporal range and number of labels - over 600 with core partner data).

AI models for computer vision have developed within the last decade, especially after the advent of ImageNet \cite{ImageNet}. They can be tuned for wildlife monitoring, and there is a growing body of work focused around camera trap data. \cite{DBLP:journals/corr/NorouzzadehNKSP17} used deep models for classifying across 48 different animal species. Another recent study from Tabak et. al. \cite{Tabak} trains a classification model over 3.7M tagged images with 28 different species collected from 5 location across the United States.

In this paper, we discuss the use of such camera trap data to extend the previous work of using deep convolutional network for fine-grained species identification with \textbf{465 classes}. We first call out the challenges we identified in using this camera trap dataset, and propose a unified setup for organizing it to build common identification models. Second, we present preliminary results from a multi-class classification model tuned on this dataset.

\section{Data}

\subsection{Sources}
Various nature conservation efforts have been collecting camera trap data over decades. \cite{TigerPopulation} and \cite{Ahumada} have used such data for studying diversity in tropical forests and the structure of communities in mammals. This data also contains additional rich information that can be used for improving the classification of the species. Hence we utilize the data from Tropical Ecology Assessment and Monitoring (TEAM) network for image classification \cite{Ahumada}. We plan to extend this data architecture to other partners in Wildlife Insights. 

The availability of low-cost cloud hosting enables de-siloing of camera trap datasets. Such data from conservation partners is combined into a common data format that is based on Open Standards for Camera Trap Data \cite{Forrester}. A subset of this data is used to train models where high confidence in expert label accuracy exists.

\subsection{Class labels}
We primarily look at the classification of mammals and aves. Our data contains 271 distinct species of mammals and 194 species of aves \ref{tab:distri}. These are tagged using the standard ontology \cite{Forrester}. We consider 5 levels - class, order, family, genus and species - coarse-grained to fine-grained.

\begin{table}[]
    \centering
    \begin{tabular}{l|cccc}
 & Order & Family & Genus & Species \\
 \hline
Mammalia & 58 & 19 & 166 & 271 \\ 
Aves & 48 & 14 & 129 & 194 \\ 
\end{tabular}
\caption{Distinct labels at different hierarchies in the ontology \label{tab:distri}}
\end{table}

\subsection{Other useful features}
Domain experts tag the images with the species ontology. Additionally, the camera captures information such as geographical location, timestamp, image bursts (i.e., a camera configuration that sets the number of images taken when triggered by an animal), along with other metadata. Such signals are useful for better animal classification, as well as monitoring their behavioral patterns like movement across different locations, nocturnal behavior and such. Hence we propose these as part of a data standard for Wildlife Insights\footnote{http://bit.ly/widataformat}.

\subsection{Challenges}
While this is a rich dataset tagged by domain experts, there are many challenges in using it to train AI models.

\begin{itemize}
    \item Falsely triggered motion sensor cameras: Since camera traps are activated by motion sensors, they may be activated by environmental factors like wind, heat waves, etc. This leads to a large number of false positive images containing no objects of interest (aka blanks), a figure that can vary between 16\% - 70\%. Blank images result in a costly human labeling effort to sift through all of the images (about 300-600 images/hour), simply to exclude them.

 \item Skewed species distribution: Over 70\% of the data is attributed to just 20 species. This skew biases the model towards the most commonly observed classes, resulting in high accuracy on common classes, but neglecting the rare classes.

\item Data quality: The camera trap data has been collected over decades. Therefore, the image quality significantly varies, and older cameras have poor resolution and less stability, resulting in blurred images. This makes it hard for the model to adapt across images with different quality.

\end{itemize}

\section{Model}
To demonstrate the use of tagged data in building intelligent models we set up a multi-class classifier for identifying species at the most fine-grained level. The model is Inception V4 pre-trained on ImageNet \cite{inception} and fine-tuned on our camera trap images. We split 2.9M images into roughly 90\% for the training set and the remaining 10\% for evaluation. We define a region as $100m^2$ using the latitude and longitude metadata associated with the images. All images from a specific region are either included in the training set or in the test set to avoid overfitting based on common background features that might be biasing the predictions to the most common species in the region. 

\section{Evaluation}
As a preliminary result, we evaluate the accuracy of the model by running the inference on the evaluation data (270K examples), that gives overall accuracy of 71.74\% @1 and 86.64\% @3.

It is worth noting that since we split the train and test sets based on $100m^2$ regions, some of the rare species may be missing from the training set altogether, and would therefore, be misclassified in evaluation. We chose the area considering the trade-off between model overfitting due to common backgrounds if the area is too small, versus losing out on rare species if the area is too large. This will be further tuned as we get additional partner datasets.

\subsection{Performance on \it{blanks}}
Since one of the use-cases is to reduce the manual burden on domain experts of sifting through images that are blanks, we evaluate the performance on the blank-class. Without any specific tuning, we get precision of 71.2\% and recall of 80.9\%. This reduces the burden on the domain experts significantly since humans can tag about 300-600 images/hour.

In follow up work, we are considering additional ways to handle such blank images like weighted loss, multi-layered model to classify blanks vs. non-blanks followed by a fine-grained classifier to improve the precision of the classifier. Using an object detector followed by a classifier is also expected to boost the precision of this task considering the inference cost versus benefit tradeoffs\cite{DBLP:journals/corr/HuangRSZKFFWSG016} .

\section{Future work}
The proposal shows the results produced by standard TensorFlow based models and has several directions for improvement. We are actively working on handling class imbalance with an intention to improve the per-class accuracy. For example, while we have very few camera trap images of the Greater Bamboo Lemur (Prolemur simus), it is a critically endangered species and needs to be given additional importance while training.

While our model design is to build an "Uber" model across data from different sites, adding geographical signals would help classify the species more accurately. For example, we can narrow our taxonomic lists to prevent mislabeling of an African elephant as an Asian Elephant if we know the image was taken in Sub-Saharan Africa (e.g. within the range of African elephants) by considering geographical features.

Given the nature of camera trap data where the images are captured in bursts across time, we expect potential gains from considering sequence models that go beyond per-image classification, but rather are trained on sequences of images. Such models would also be useful for classifying images that are falsely triggered and do not contain any objects of interest.

The model is integrated into the Wildlife Insights\footnote{https://wildlifeinsights.org} platform, a joint effort of seven leading conservation organizations to quantitatively measure biodiversity data and offer more timely geospatial data products to aid in decision-making. Empowered with this new dataset from in-situ sensors, researchers will be able to make more informed conservation recommendations in a time when widespread biodiversity decline is an urgent crisis.

\begin{acks}
We thank our collaborators Jonathan Huang, Christine Kaeser-Chen, Wildlife Insights partners, Katherine Chou, Sara Beery, Rebecca Moore, Karin Tuxen-Bettman for their contribution to the discussion. 
\end{acks}

\nocite{classimbalance}
\nocite{iWildCam}

\bibliographystyle{ACM-Reference-Format}
\bibliography{wildnet}


\begin{thebibliography}{10}


\ifx \showCODEN    \undefined \def \showCODEN     #1{\unskip}     \fi
\ifx \showDOI      \undefined \def \showDOI       #1{#1}\fi
\ifx \showISBNx    \undefined \def \showISBNx     #1{\unskip}     \fi
\ifx \showISBNxiii \undefined \def \showISBNxiii  #1{\unskip}     \fi
\ifx \showISSN     \undefined \def \showISSN      #1{\unskip}     \fi
\ifx \showLCCN     \undefined \def \showLCCN      #1{\unskip}     \fi
\ifx \shownote     \undefined \def \shownote      #1{#1}          \fi
\ifx \showarticletitle \undefined \def \showarticletitle #1{#1}   \fi
\ifx \showURL      \undefined \def \showURL       {\relax}        \fi
\providecommand\bibfield[2]{#2}
\providecommand\bibinfo[2]{#2}
\providecommand\natexlab[1]{#1}
\providecommand\showeprint[2][]{arXiv:#2}

\bibitem[\protect\citeauthoryear{Ahumada, Silva, Gajapersad, Hallam, Hurtado,
  Martin, McWilliam, Mugerwa, O'Brien, Rovero, Sheil, Spironello, Winarni, and
  Andelman}{Ahumada et~al\mbox{.}}{2011}]%
        {Ahumada}
\bibfield{author}{\bibinfo{person}{Jorge Ahumada}, \bibinfo{person}{Carlos
  Silva}, \bibinfo{person}{Krisna Gajapersad}, \bibinfo{person}{Christopher
  Hallam}, \bibinfo{person}{Johanna Hurtado}, \bibinfo{person}{Emanuel Martin},
  \bibinfo{person}{Alex McWilliam}, \bibinfo{person}{Badru Mugerwa},
  \bibinfo{person}{Timothy O'Brien}, \bibinfo{person}{Francesco Rovero},
  \bibinfo{person}{Douglas Sheil}, \bibinfo{person}{Wilson Spironello},
  \bibinfo{person}{Nurul Winarni}, {and} \bibinfo{person}{Sandy Andelman}.}
  \bibinfo{year}{2011}\natexlab{}.
\newblock \showarticletitle{Community structure and diversity of tropical
  forest mammals: Data from a global camera trap network}.
\newblock \bibinfo{journal}{\emph{Philosophical transactions of the Royal
  Society of London. Series B, Biological sciences}}  \bibinfo{volume}{366}
  (\bibinfo{date}{09} \bibinfo{year}{2011}), \bibinfo{pages}{2703--11}.
\newblock
\urldef\tempurl%
\url{https://doi.org/10.1098/rstb.2011.0115}
\showDOI{\tempurl}


\bibitem[\protect\citeauthoryear{Beery, Horn, {Mac Aodha}, and Perona}{Beery
  et~al\mbox{.}}{2019}]%
        {iWildCam}
\bibfield{author}{\bibinfo{person}{Sara Beery}, \bibinfo{person}{Grant~Van
  Horn}, \bibinfo{person}{Oisin {Mac Aodha}}, {and} \bibinfo{person}{Pietro
  Perona}.} \bibinfo{year}{2019}\natexlab{}.
\newblock \showarticletitle{The iWildCam 2018 Challenge Dataset}.
\newblock \bibinfo{journal}{\emph{CoRR}}  \bibinfo{volume}{abs/1904.05986}
  (\bibinfo{year}{2019}).
\newblock
\showeprint[arxiv]{1904.05986}
\urldef\tempurl%
\url{http://arxiv.org/abs/1904.05986}
\showURL{%
\tempurl}


\bibitem[\protect\citeauthoryear{Buda, Maki, and Mazurowski}{Buda
  et~al\mbox{.}}{2017}]%
        {classimbalance}
\bibfield{author}{\bibinfo{person}{Mateusz Buda}, \bibinfo{person}{Atsuto
  Maki}, {and} \bibinfo{person}{Maciej~A. Mazurowski}.}
  \bibinfo{year}{2017}\natexlab{}.
\newblock \showarticletitle{A systematic study of the class imbalance problem
  in convolutional neural networks}.
\newblock \bibinfo{journal}{\emph{CoRR}}  \bibinfo{volume}{abs/1710.05381}
  (\bibinfo{year}{2017}).
\newblock
\showeprint[arxiv]{1710.05381}
\urldef\tempurl%
\url{http://arxiv.org/abs/1710.05381}
\showURL{%
\tempurl}


\bibitem[\protect\citeauthoryear{Forrester, O'Brien, Fegraus, A Jansen,
  Palmer, Kays, Ahumada, Stern, and McShea}{Forrester et~al\mbox{.}}{2016}]%
        {Forrester}
\bibfield{author}{\bibinfo{person}{Tavis Forrester}, \bibinfo{person}{Tim
  O'Brien}, \bibinfo{person}{Eric Fegraus}, \bibinfo{person}{Patrick
  A Jansen}, \bibinfo{person}{Jonathan Palmer}, \bibinfo{person}{Roland Kays},
  \bibinfo{person}{Jorge Ahumada}, \bibinfo{person}{Beth Stern}, {and}
  \bibinfo{person}{William McShea}.} \bibinfo{year}{2016}\natexlab{}.
\newblock \showarticletitle{An Open Standard for Camera Trap Data}.
\newblock   \bibinfo{volume}{4} (\bibinfo{year}{2016}).
\newblock
\urldef\tempurl%
\url{https://doi.org/10.3897/BDJ.4.e10197}
\showDOI{\tempurl}


\bibitem[\protect\citeauthoryear{Huang, Rathod, Sun, Zhu, Korattikara, Fathi,
  Fischer, Wojna, Song, Guadarrama, and Murphy}{Huang et~al\mbox{.}}{2016}]%
        {DBLP:journals/corr/HuangRSZKFFWSG016}
\bibfield{author}{\bibinfo{person}{Jonathan Huang}, \bibinfo{person}{Vivek
  Rathod}, \bibinfo{person}{Chen Sun}, \bibinfo{person}{Menglong Zhu},
  \bibinfo{person}{Anoop Korattikara}, \bibinfo{person}{Alireza Fathi},
  \bibinfo{person}{Ian Fischer}, \bibinfo{person}{Zbigniew Wojna},
  \bibinfo{person}{Yang Song}, \bibinfo{person}{Sergio Guadarrama}, {and}
  \bibinfo{person}{Kevin Murphy}.} \bibinfo{year}{2016}\natexlab{}.
\newblock \showarticletitle{Speed/accuracy trade-offs for modern convolutional
  object detectors}.
\newblock \bibinfo{journal}{\emph{CoRR}}  \bibinfo{volume}{abs/1611.10012}
  (\bibinfo{year}{2016}).
\newblock
\showeprint[arxiv]{1611.10012}
\urldef\tempurl%
\url{http://arxiv.org/abs/1611.10012}
\showURL{%
\tempurl}


\bibitem[\protect\citeauthoryear{Norouzzadeh, Nguyen, Kosmala, Swanson, Packer,
  and Clune}{Norouzzadeh et~al\mbox{.}}{2017}]%
        {DBLP:journals/corr/NorouzzadehNKSP17}
\bibfield{author}{\bibinfo{person}{Mohammad~Sadegh Norouzzadeh},
  \bibinfo{person}{Anh Nguyen}, \bibinfo{person}{Margaret Kosmala},
  \bibinfo{person}{Ali Swanson}, \bibinfo{person}{Craig Packer}, {and}
  \bibinfo{person}{Jeff Clune}.} \bibinfo{year}{2017}\natexlab{}.
\newblock \showarticletitle{Automatically identifying wild animals in camera
  trap images with deep learning}.
\newblock \bibinfo{journal}{\emph{CoRR}}  \bibinfo{volume}{abs/1703.05830}
  (\bibinfo{year}{2017}).
\newblock
\showeprint[arxiv]{1703.05830}
\urldef\tempurl%
\url{http://arxiv.org/abs/1703.05830}
\showURL{%
\tempurl}


\bibitem[\protect\citeauthoryear{Russakovsky, Deng, Su, Krause, Satheesh, Ma,
  Huang, Karpathy, Khosla, Bernstein, Berg, and Fei-Fei}{Russakovsky
  et~al\mbox{.}}{2015}]%
        {ImageNet}
\bibfield{author}{\bibinfo{person}{Olga Russakovsky}, \bibinfo{person}{Jia
  Deng}, \bibinfo{person}{Hao Su}, \bibinfo{person}{Jonathan Krause},
  \bibinfo{person}{Sanjeev Satheesh}, \bibinfo{person}{Sean Ma},
  \bibinfo{person}{Zhiheng Huang}, \bibinfo{person}{Andrej Karpathy},
  \bibinfo{person}{Aditya Khosla}, \bibinfo{person}{Michael Bernstein},
  \bibinfo{person}{Alexander~C. Berg}, {and} \bibinfo{person}{Li Fei-Fei}.}
  \bibinfo{year}{2015}\natexlab{}.
\newblock \showarticletitle{{ImageNet Large Scale Visual Recognition
  Challenge}}.
\newblock \bibinfo{journal}{\emph{International Journal of Computer Vision
  (IJCV)}} \bibinfo{volume}{115}, \bibinfo{number}{3} (\bibinfo{year}{2015}),
  \bibinfo{pages}{211--252}.
\newblock
\urldef\tempurl%
\url{https://doi.org/10.1007/s11263-015-0816-y}
\showDOI{\tempurl}


\bibitem[\protect\citeauthoryear{Sharma and Jhala}{Sharma and Jhala}{2010}]%
        {TigerPopulation}
\bibfield{author}{\bibinfo{person}{Rishi Sharma} {and}
  \bibinfo{person}{Yadvendradev Jhala}.} \bibinfo{year}{2010}\natexlab{}.
\newblock \showarticletitle{Monitoring tiger populations using intensive search
  in a capture-recapture framework}.
\newblock \bibinfo{journal}{\emph{Population Ecology}}  \bibinfo{volume}{53}
  (\bibinfo{date}{04} \bibinfo{year}{2010}), \bibinfo{pages}{373--381}.
\newblock
\urldef\tempurl%
\url{https://doi.org/10.1007/s10144-010-0230-9}
\showDOI{\tempurl}


\bibitem[\protect\citeauthoryear{Szegedy, Ioffe, Vanhoucke, and Alemi}{Szegedy
  et~al\mbox{.}}{2016}]%
        {inception}
\bibfield{author}{\bibinfo{person}{Christian Szegedy}, \bibinfo{person}{Sergey
  Ioffe}, \bibinfo{person}{Vincent Vanhoucke}, {and} \bibinfo{person}{Alex~A.
  Alemi}.} \bibinfo{year}{2016}\natexlab{}.
\newblock \showarticletitle{Inception-v4, Inception-ResNet and the Impact of
  Residual Connections on Learning}. In \bibinfo{booktitle}{\emph{ICLR 2016
  Workshop}}.
\newblock
\urldef\tempurl%
\url{https://arxiv.org/abs/1602.07261}
\showURL{%
\tempurl}


\bibitem[\protect\citeauthoryear{Tabak, Norouzzadeh, Wolfson, Sweeney,
  Vercauteren, Snow, M~Halseth, A~Di~Salvo, Lewis, White, Teton, Beasley,
  Schlichting, Boughton, Wight, S~Newkirk, S~Ivan, A~Odell, K~Brook, and
  Miller}{Tabak et~al\mbox{.}}{2018}]%
        {Tabak}
\bibfield{author}{\bibinfo{person}{Michael Tabak},
  \bibinfo{person}{Mohammad~Sadegh Norouzzadeh}, \bibinfo{person}{David
  Wolfson}, \bibinfo{person}{Steven Sweeney}, \bibinfo{person}{Kurt
  Vercauteren}, \bibinfo{person}{Nathan Snow}, \bibinfo{person}{Joseph
  M~Halseth}, \bibinfo{person}{Paul A~Di~Salvo}, \bibinfo{person}{Jesse Lewis},
  \bibinfo{person}{Michael White}, \bibinfo{person}{Ben Teton},
  \bibinfo{person}{James Beasley}, \bibinfo{person}{Peter Schlichting},
  \bibinfo{person}{Raoul Boughton}, \bibinfo{person}{Bethany Wight},
  \bibinfo{person}{Eric S~Newkirk}, \bibinfo{person}{Jacob S~Ivan},
  \bibinfo{person}{Eric A~Odell}, \bibinfo{person}{Ryan K~Brook}, {and}
  \bibinfo{person}{Ryan Miller}.} \bibinfo{year}{2018}\natexlab{}.
\newblock \bibinfo{title}{Machine learning to classify animal species in camera
  trap images: applications in ecology}.
\newblock
\newblock
\urldef\tempurl%
\url{https://doi.org/10.1101/346809}
\showDOI{\tempurl}


\end{thebibliography}

\end{document}